\pdfoutput=1

\documentclass[11pt]{article}

\usepackage[]{EMNLP2022}
\usepackage{enumitem}
\usepackage[T1]{fontenc}
\usepackage[utf8]{inputenc}
\usepackage{comment}

\usepackage{microtype}
\usepackage{inconsolata}

\usepackage{xcolor}
\definecolor{Green}{rgb}{0.0, 0.5, 0.0}
\definecolor{Amethyst}{rgb}{0.6, 0.4, 0.8}

\usepackage{times}
\usepackage{latexsym}

\usepackage{multicol}
\usepackage{multirow}
\usepackage{graphicx}
\usepackage{verbatim}
\usepackage{makecell}
\usepackage{braket}
\usepackage{hhline}
\usepackage{amsmath,bm,amssymb}
\usepackage[linesnumbered,ruled]{algorithm2e}

\title{Learning to Revise References\\for Faithful Summarization} 

\author{

\quad \textbf{Griffin Adams$^{1}$}\thanks{\ \ This project was completed during an NLP research internship with Amazon Comprehend Medical.}
\quad \textbf{Han-Chin Shing$^{2}$}
\quad Qing Sun$^{2}$ \\
\quad \textbf{Christopher Winestock}$^{2}$
\quad \textbf{Kathleen McKeown}$^{1,2}$
\quad \textbf{No\'emie Elhadad}$^{1}$ \\
\texttt{\{griffin.adams, noemie.elhadad\}@columbia.edu} \\
\texttt{\{hanchins, qinsun, winestock, mckeownk\}@amazon.com} \\
\quad $^{1}$Columbia University, New York, NY \quad $^{2}$Amazon AWS AI, Seattle, WA
}

\begin{document}
\maketitle
\begin{abstract}

In real-world scenarios with naturally occurring datasets, reference summaries are noisy and may contain information that cannot be inferred from the source text. On large news corpora, removing low quality samples has been shown to reduce model hallucinations. Yet, for smaller, and/or noisier corpora, filtering is detrimental to performance. To improve reference quality while retaining all data, we propose a new approach: to selectively re-write \emph{un}supported reference sentences to better reflect source data. We automatically generate a synthetic dataset of positive and negative revisions by corrupting supported sentences and learn to revise reference sentences with contrastive learning. The intensity of revisions is treated as a controllable attribute so that, at inference, diverse candidates can be over-generated-then-rescored to balance faithfulness and abstraction. To test our methods, we extract noisy references from publicly available MIMIC-III discharge summaries for the task of hospital-course summarization, and vary the data on which models are trained. According to metrics and human evaluation, models trained on revised clinical references are much more faithful, informative, and fluent than models trained on original or filtered data.




\end{abstract}

\section{Introduction}

\begin{figure}[t]\centering
\includegraphics[width= \linewidth]{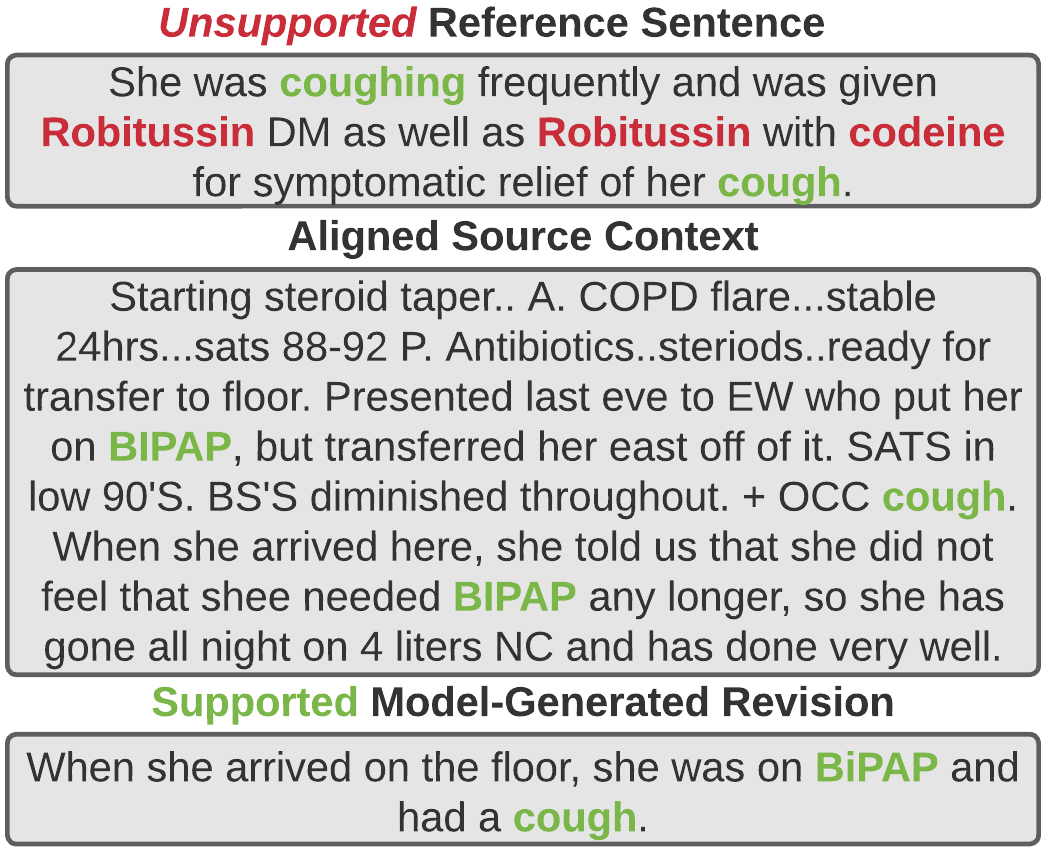}
\caption{Example of a revised reference sentence. \textbf{\textcolor{red}{Robitussin}} and \textbf{\textcolor{red}{codeine}} are edited out of the sentence, while \textbf{\textcolor{Green}{cough}} is correctly kept and a new supported entity \textbf{\textcolor{Green}{BiPAP}} is added. The model is trained on synthetic data to reconstruct well-supported sentences based on context and diverse model-generated hallucinations.} 
\label{fig:revision-example}
\centering
\end{figure}

The tendency of abstractive systems to produce unfaithful summaries is well-studied \citep{maynez-etal-2020-faithfulness}, yet less attention is paid to the role of the data on which the models are trained. This is problematic for two reasons: \textbf{(1)} many corpora are naturally occurring--not created for training models--and, as such, are noisy \citep{kryscinski2019critical} and without ``inherent quality guarantees'' \citep{bommasani-cardie-2020-intrinsic}; \textbf{(2)} noisy data is detrimental to training faithful models \citep{duvsek2019semantic}.


A common approach to deal with training noise is \emph{filtering}: to identify and ignore low quality text at the reference \citep{kang-hashimoto-2020-improved, matsumaru-etal-2020-improving, nan2021entity, narayan2021planning} or span level \citep{goyal2021annotating}. Yet, these methods largely work because they are applied to clean, large-scale corpora. For instance, after removing references with ``entity hallucinations'', \citet{nan2021entity} still have 855k (92\% of the original) training examples for Newsroom, 286k (99\%) for CNN/DM, 135k (66\%) for Xsum. 


We consider a noisier, lower resource setting (clinical summarization) and propose a new approach: to revise--not remove--noisy reference content. First, we align each reference sentence to 1-5 sentences in the source text and classify it as \emph{supported} (to be left alone) or \emph{unsupported} (to be revised). Our objective is to revise all unsupported reference sentences in such a way that retains faithful content, removes unfaithful content, and, as needed to preserve length, adds relevant context. An example output is shown in Figure \ref{fig:revision-example}. In a coherent sentence, the model removes unsupported entities (\textbf{\textcolor{red}{Robitussin}}, \textbf{\textcolor{red}{codeine}}) and introduces a twice mentioned concept from the context (\textbf{\textcolor{Green}{BiPAP}}).

To learn this revision task, we need examples of supported and unsupported reference sentences for the same context. Without directly observing it, we generate synthetic data. At a high-level, we take each supported sentence, corrupt it to form a diverse set of unsupported alternatives, and use this mix of real and synthetic data to create examples of (un)faithful revisions for contrastive learning.


As a test case, we consider a task of real-world significance--summarizing a hospital admission--and extract a corpus from a noisy source--notes from the Electronic Health Record (EHR). We experiment with the publicly available MIMIC-III dataset \citep{johnson2016mimic}. As in \citet{adams-etal-2021-whats}, we treat the Brief Hospital Course (BHC) section of the discharge summary as a reference summary and all notes prior to discharge as source. Data coverage is a huge issue as only 60\% of reference summary entities can be found in the source.


The contributions of this work are: \textbf{(1)} Proposing a new method to address variable reference quality: reference revision, which, as a data pre-processing step, is model agnostic and complementary to other faithfulness approaches; \textbf{(2)} Showing that training on revised references can improve faithfulness while also improving informativeness and fluency; \textbf{(3)} Providing code\footnote{\url{ https://github.com/amazon-research/summary-reference-revision}}, pre-processed datasets, and models for alignment, corruption, revision, post-hoc editing, and generation of clinical summaries\footnote{The datasets are accessible via PhysioNet \citep{PhysioNet} and models via HuggingFace \citep{wolf-etal-2020-transformers}.  Please refer to our GitHub README for further details.}; \textbf{(4)} Beyond its primary use case of data pre-processing, demonstrating that reference revision can have standalone value: as a post-hoc editor and a pre-training objective for faithfulness.

\section*{Related Work}

\paragraph{Faithfulness.} Efforts to address faithfulness have focused on smarter models \citep{huang-etal-2020-knowledge}, more suitable metrics \citep{durmus-etal-2020-feqa}, content-plan editing \citep{narayan2021planning}, and post-hoc interventions: ranking \citep{falke-etal-2019-ranking} and editing \citep{cao-etal-2020-factual,dong-etal-2020-factcorrect,zhu-etal-2021-enhancing}. Synthetic errors \citep{kryscinski-etal-2020-evaluating} are useful for optimizing \citep{cao2021cliff} and evaluating \citep{goyal-durrett-2020-evaluating} faithfulness, yet are best supplemented with fine-grained annotations \citep{goyal2021annotating}.



The impact of training data noise on faithfulness is less studied. The most common proposal is to identify low quality samples--with entailment \citep{matsumaru-etal-2020-improving, goyal2021annotating} or entity overlap \citep{nan2021entity, narayan2021planning}--and drop them. These filtering methods tend to improve faithfulness yet can degrade informativeness. \citet{kang-hashimoto-2020-improved} address the data hunger issue by first training on all samples before implementing Loss Truncation--ignoring high log loss datapoints. This is effective on a relatively clean Gigaword corpus yet untested on noisier corpora in which hallucination behavior from the unfiltered data may be difficult to unlearn. Filtering can be ``insufficient to train a competitive model'' when high-quality data is limited \citep{filippova-2020-controlled}. Our proposed method takes advantage of all available data while seeking to redress the underlying issue.





\begin{figure*}[t]
\centering
\includegraphics[width=\linewidth]{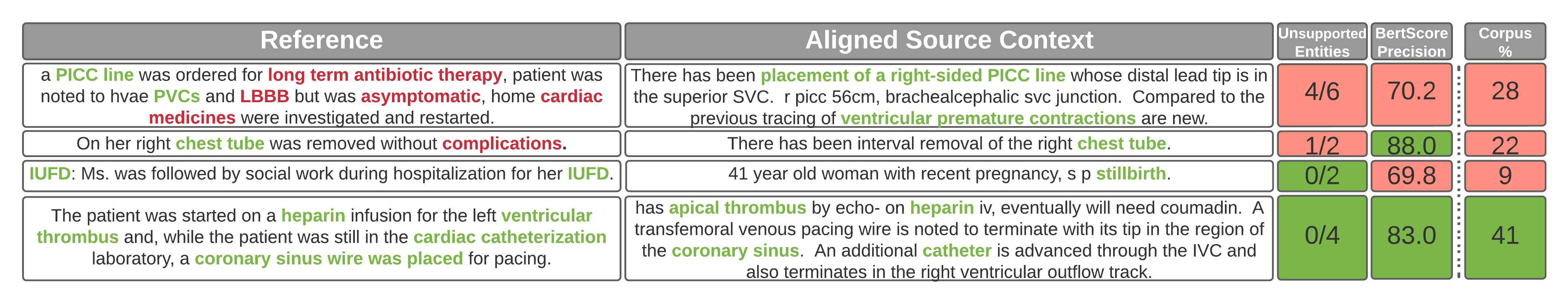}
\caption{From top to bottom: examples of source-reference alignments which fail to meet both criteria for supportedness (only supported entities and a high BERTScore precision) (28\% of corpus reference sentences), just entity overlap (22\%), just low BERTScore (9\%), and the remaining which meet both (41\%). Only the last reference sentence is classified as \emph{supported} and, for this paper, treated as a gold standard when training revisions. } 
\label{fig:quality}
\end{figure*}

\paragraph{Clinical Summarization.} Faithfulness is less studied for clinical summarization because most proposed methods are extractive \citep{pivovarov2015automated,moen2016comparison,alsentzer2018extractive}. Abstractive approaches tend to focus on finer temporal granularities, e.g., synthesizing a single radiology report \citep{macavaney2019ontology, sotudeh2020attend, zhang-etal-2020-optimizing} and doctor-patient conversations \citep{krishna2020generating,joshi2020dr, zhang2021leveraging}.  

Most similar, \citet{shing2021towards} survey extract-then-abstract approaches to section-specific discharge summarization on MIMIC-III. They measure factuality with entity overlap and remove poorly supported references. When analyzing a proprietary EHR-derived, hospital-course summarization corpus, \citet{adams-etal-2021-whats} perform oracle extractive analysis and confirm, as we do, that EHR-derived summary references are highly noisy. 


\begin{table}[t]
\centering
\footnotesize
\begin{tabular}{clc}
\hline
& \textbf{Statistic} & \textbf{Value}\\
\hline
\multirow{2}{*}{Global}
& Notes & 1.38M \\
& Unique Patients & 47,553 \\
\hline
\multirow{6}{*}{\makecell{Per \\ Admission\\ (Avg. \#s)}}
& Notes & 29 \\
& Note types & 3.3 \\
& Source sentences & 703 \\
& Source tokens & 7,553 \\
& Reference sentences & 23.5 \\
& Reference tokens & 370 \\
\multirow{3}{*}{\makecell{Extractive \\ Analysis}}
& Coverage & 46.0 \\
& Density & 1.2 \\
& Compression Ratio & 20 \\\hline
\end{tabular}
\caption{Hospital-Admission Summarization Dataset.}\label{tab:data}
\end{table}

\section{Data}

The publicly available MIMIC-III dataset contains de-identified clinical records from patients admitted to Beth Israel Deaconess Medical Center \citep{johnson2016mimic}. Hospital-admission summarization is a challenging task: in at most a paragraph, a summary must discuss what happened to the patient during the admission, why it happened, and what needs to happen next. To create a dataset for this task, we treat the Brief Hospital Course section of the discharge summary as a reference and all notes authored during the patient's stay as the source. Table \ref{tab:data} shows that source documents are long (on average, 7.5K tokens and 703 sentences) and, while references are also long (370 tokens on average and 23.5 sentences), there is a high degree of word-level compression ($\sim$20x).  Coverage and density metrics reveal high levels of abstraction\footnote{Please refer to \citet{grusky2018newsroom} for details on extractive analysis, including formulas for density and coverage.}.




 \href{https://aws.amazon.com/comprehend/medical/}{Amazon Comprehend Medical} is used to extract entities of the following semantic types: diagnoses (using the ICD-10 classification), medications (RxNorm), procedures, treatments, and tests. To determine whether or not an entity is supported, we compute a similarity score for all mention pairs based on a combination of ontological (RxNorm/ICD-10 codes) and lexical overlap (embedding and exact match) (see Appendix \ref{app:entity-merging}).



\section{Building Source-Reference Alignments} \label{sec:alignment}

We link each reference sentence to a subset of source sentences to identify the minimal context necessary for revision and determine which sentences need to be revised. We select no more than five sentences because \citet{lebanoff-etal-2019-scoring} find that reference sentences tend to reflect content from very few source sentences. We follow their approach to greedily select sentences with high ROUGE \citep{lin2004rouge} overlap and minimize redundancy by removing covered tokens after each step. Yet, given high levels of abstraction (abbreviations \citep{adams2020zero}, misspellings), we increase semantic coverage by replacing ROUGE with BERTScore precision \citep{zhang2019bertscore} and adding additional sentences, as needed, to cover all supported entities. Please refer to Appendix \ref{app:alignment} for intuition, notation, and an example.

\paragraph{Classifying References.} We treat reference sentences with $0$ unsupported entities and a BERTScore precision with respect with its aligned source evidence of $\geq 0.75$ as supported. The remaining are unsupported. 417,318 (41\%) reference sentences qualify as supported and the remaining 595,300 (59\%) unsupported: 47\% fail both thresholds (280,839), 38\% have hallucination(s) with a high BERTScore (225,423), and 15\% have no hallucinations but a low BERTScore (88,189). Figure \ref{fig:quality} reveals why both BERTScore and entity overlap are needed to identify full coverage. The first sentence has unsupported entities and poor embedding-based scores. The second is semantically similar yet missing a critical concept (\textcolor{red}{complications}) which cannot be inferred from the context. The third has full entity coverage (\textcolor{Green}{IUFD} is a term for \textcolor{Green}{stillbirth}) yet BERTScore is low because there is no mention of social work. Only the final sentence is covered by both metrics and treated as supported.

\section{Learning to Revise Unsupported Text} \label{sec:revise-pipeline}

The goal is to re-write these \emph{un}supported reference sentences such that they are supported, i.e., covered by the source notes. To learn this revision task without a gold standard ground-truth, we take each supported reference sentence, inject noise to create unsupported, yet realistic, alternatives (\S \ref{sec:perturb}), and then use this mix of real and synthetic data to create a supervised set of positive and negative revisions to support a contrastive learning objective (\S \ref{sec:reviser})\footnote{We do not rely on \emph{un}supported sentences during training because they are set aside for inference. To use them, we would need to synthetically construct supported alternatives, which is not possible without first knowing how to revise.}.

\subsection{Generating \textit{Synthetic} Hallucinations} \label{sec:perturb}

\paragraph{(D)esiderata.} Based on Figure \ref{fig:quality}, unsupported sentences look normal (unfaithful only in the context of available data) \textbf{(D1)}; contain many hallucinated entities \textbf{(D2)}; exhibit a wide range of semantic divergence from the aligned source context \textbf{(D3)}; and in spite of clear differences, are topically similar to aligned context \textbf{(D4)}. To illustrate \textbf{D3}, we note that the second sentence could be revised by simply removing the bigram ``without complications'', yet the first and third sentences require more substantial re-writing to remove unfaithful content while still remaining informative and coherent.

\paragraph{High-Level.} The simplest way to construct, and control for, hallucinations is to perform entity swaps \citep{kryscinski-etal-2020-evaluating,zhao2020reducing, zhang2021fine,chen2021improving}. Yet, this can produce disfluent text which is easily detectable \citep{goyal2021annotating}. In contrast, generating from a LLM produces fluent \citep{zhou2020detecting}, more diverse text \citep{cao2021cliff}, yet without as much control over hallucinations. Given our desiderata, we combine entity swaps \textbf{(D2)} into a generative framework \textbf{(D1, D3)} and incorporate a set of topical entities to avoid excessive semantic drift \textbf{(D4)}. Our proposed method is called \textbf{ReDRESS}: \textbf{re}ference \textbf{d}ist\textbf{r}actor \textbf{e}ntity \textbf{s}et \textbf{s}wapping.

\paragraph{Training Objective.} The \textbf{ReDRESS} backbone is a BART encoder-decoder model \citep{lewis-etal-2020-bart}. BART is trained as a denoising autoencoder to reconstruct a corrupted sentence: $p(s|f(s))$, where $f$ is an arbitrary noise function(s). Across natural language tasks, the BART authors find span deletion to be the most effective noise. \textbf{ReDRESS} also uses span deletion but adds an extra noise function: entity swaps. Specifically, for each sentence $s$, we extract a set of topically related entities $\bm{e_s}$ and then exchange entities between $s$ and $\bm{e_s}$. Let us denote span deletion as $f$ and the swap transformation as $g(s,\bm{e_s}, k) \rightarrow {\bm{e_s}}_{-k}^{+k}, s_{-k}^{+k}$, where $k$ represents the number of entities exchanged. To vary the level of corruption (\textbf{D3}), we sample a different $k$ for each example such that, on average, half of the entity mentions in $s$ are swapped out for entities in $\bm{e_s}$. The final pre-training objective is $p(s|k,{\bm{e_s}}_{-k}^{+k}, f(s)_{-k}^{+k})$. $k$ is represented by a special token and added to the input to make entity swaps a controllable aspect of generation. Each component ($k$, ${\bm{e_s}}_{-k}^{+k}$, $f(s)_{-k}^{+k}$) is separated by a special \texttt{<sep>} token and passed to the BART encoder. During training, the decoder learns to reconstruct $s$ by re-writing $f(s)_{-k}^{+k}$ such that it reverses the $k$ entity swaps performed between $\bm{e_s}$ and $s$ and fills in a deleted text span\footnote{$\bm{e_s}$ is provided as input yet is not part of the target output. In other words, the output of the model is a natural sentence.}.


\begin{figure*}[t]
\centering
\includegraphics[width=\linewidth]{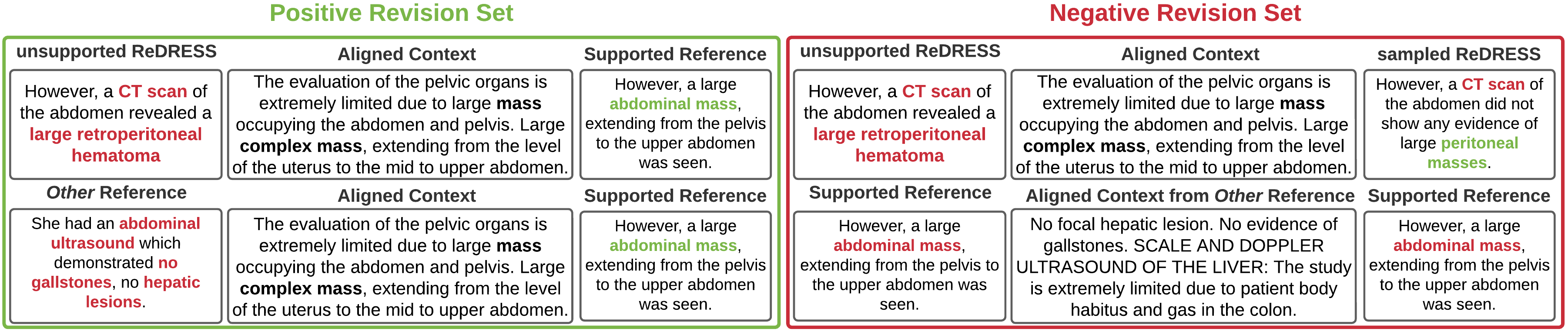}
\caption{Synthetic \textcolor{Green}{positive} and \textcolor{red}{negative} sets for revision training. The encoder input is the concatenation of the input (first box) and source context (second box), while the (un)faithful revision target is the third. Entities from inputs and targets are colored as \textcolor{red}{un}\textcolor{Green}{supported} \textbf{\emph{relative}} to the provided context. \textit{Sampled ReDRESS} is a randomly sampled synthetic hallucination, while \textit{Unsupported ReDRESS} is the sample most unsupported by \textit{Aligned Context}. Figure \ref{fig:reviser-architecture} (Appendix \ref{app:reviser-detail}) visualizes how this data is obtained from ReDRESS and within-example misalignments.}
\label{fig:revise-train-example}
\end{figure*}

\paragraph{Inference.} To use \textbf{ReDRESS} to generate a plausible, corrupted version of a sentence $s$, we apply $f$ to $s$ and sample $k$ to apply $g$ to both $s$ and its distractor set $\bm{e_s}$. Two key \textbf{m}odifications are introduced to discourage the model from reconstructing $s$: \textbf{(m1)} entities removed from $s$ are not added to $\bm{e_s}$, and \textbf{(m2)} $k$ swaps are implemented, yet the model is provided $k+1$ as the swap code to trick the model into performing an additional swap than is required for reconstruction. Using the notation above, \textbf{ReDRESS} generates $(k+1, {\bm{e_s}}_{-k}, f(s)_{-k}^{+k}) \rightarrow \hat{s}$ using standard beam search decoding. Without access to the original entities from $s$ in $\bm{e_s}$ \textbf{(m1)}, the model looks for plausible, \textit{hallucinated} alternatives. In Appendix \ref{app:perturber-train-details}, we show that \textbf{(m1, m2)} increase the diversity of synthetic hallucinations $\hat{s}$ to mirror the variable level of coverage observed in the data.

\paragraph{Implementation Details.} We train \textbf{ReDRESS} from \texttt{bart-base} on a large set of unlabeled sentences extracted from MIMIC-III discharge summaries. To find the distractor set $\bm{e_s}$ specific to each $s$, we first retrieve the sentences most similar to $s$ (using BioSentVec \citep{chen2019biosentvec} embeddings to create a Faiss index \citep{JDH17}). The first 25 unique concepts extracted from the set of nearest neighbors form the distractor set $\bm{e_s}$ for $s$.

\subsection{Learning a Revision Model} \label{sec:reviser}

We apply \textbf{ReDRESS} to the set of supported reference summary sentences to generate positive and negative revision examples for contrastive learning.

\paragraph{Notation.} Let $\bm{r}$ represent a reference sentence with aligned source sentences $\bm{S}$. $\bm{\hat{r}_n}$ is a corrupted version of $\bm{r}$ generated by \textbf{ReDRESS} with random seed $\bm{n}$. Given $N$ diverse outputs, each generated from its own sampled set of corruptions, $\bm{\hat{r}_u}$ is the most \textbf{u}nsupported (lowest BERTScore precision). 


\paragraph{Training.} The input to the \textbf{reviser} model is the concatenation of a noisy input and aligned source context, while the output is a (un)faithful revision. We rely on \textbf{ReDRESS} to generate noisy inputs \textit{and} unfaithful outputs. \textbf{ReDRESS} hallucinations require moderate levels of revision to reconstruct the original supported input. This makes sense for most of the \textit{observed} unsupported sentences. Yet, sometimes, a reference sentence is almost entirely unsupported. In such cases, the model should effectively learn to ignore it and just summarize the aligned context. To simulate this more drastic scenario, we also retrieve a \textbf{random} reference sentence ($\bm{r^*}$), and its aligned source ($\bm{S^*}$), from the same example. Our ablation study in Table \ref{tab:results-ablation} shows that both sources of hallucinated content (\textbf{ReDRESS}-generated and random mis-alignments) are complementary, necessary to achieve the best downstream summary results.

Using the notation above, as tuples of format (input, context, target), the positive set is: $(\bm{\hat{r}_u}, \bm{S}, \bm{r})$ and $(\bm{r^*}, \bm{S}, \bm{r})$. The negative set is: $(\bm{\hat{r}_u}, \bm{S}, \bm{\hat{r}_{c \in N}})$ and $(\bm{r}, \bm{S^*}, \bm{r})$, where $\bm{\hat{r}_{n \in N}}$ is a randomly selected corruption. In other words, for the positive set, we learn to generate a supported reference $\bm{r}$ from its aligned context ($\bm{S}$) and either a \textbf{ReDRESS} output ($\bm{\hat{r}_u}$) or another reference sentence ($\bm{r^*}$) as the synthetic, unsupported input. For the negatives, we discourage the model from generating (1) a synthetic hallucination ($\bm{\hat{r}_{n \in N}}$), and (2) itself if unsupported. Figure \ref{fig:revise-train-example} shows an example from the training data.

As in \texttt{ConSeq} \citep{nan2021improving}, we optimize the likelihood of positives ($\bm{Z^+}$) and unlikelihood \citep{welleck2019neural} of negatives ($\bm{Z^-}$):

\begin{align}
\label{eq:revision}
\begin{split}
  \mathcal{L}_{contrast} = \mathbb{E}_{\bm{Z^+}} log (p_\theta(\bm{r_{out}}|\bm{r_{in}}, \bm{S})) -\\
  \mathbb{E}_{\bm{Z^-}} log (1 - p_\theta(\bm{r_{out}}|\bm{r_{in}}, \bm{S}))
  \end{split}
\end{align}

$\bm{r_{in}}$ stands for the noisy reference input and $\bm{r_{out}}$ the revision target (positive or negative). We concatenate $r_{in}$ and $S$ with a special \texttt{<SEP>} token as the input, in addition to two key revision codes ($input_{frac}$ and $source_{frac}$) which we discuss next.

\paragraph{Controlling Revision Intensity.} Some sentences require minimal edits to be fully supported while others require major edits. This variance is difficult to learn without an explicit control for it. Qualitatively, we noticed a tendency of the revision model to over-edit mostly supported sentences and under-edit highly unsupported content. Given this, we introduce the notion of revision intensity, parameterized by the fraction of words in the revision copied from the input ($input_{frac} = \frac{|\bm{r_{out}} \cap \bm{r_{in}}|}{|\bm{r_{out}}|}$), and the fraction copied from the aligned context ($source_{frac} = \frac{|\bm{r_{out}} \cap \bm{S}|}{|\bm{r_{out}}|}$). Intense revisions tend to require a larger lexical shift from input to source: a low $input_{frac}$ and a high $source_{frac}$. During training, we bin the fractions into deciles and include them as style codes prefixed to the encoder. Our ablation study in Table \ref{tab:results-ablation} shows that controlling the intensity of revisions to support diverse candidate generation, followed by re-scoring, has a massive impact on downstream summaries.

\paragraph{Inference.}  We apply the trained reviser model to all unsupported reference sentences in the summarization training dataset. In particular, we concatenate each unsupported sentence as $r_{in}$ to its aligned context $S$ for beam search decoding. For this set of sentences, the desired revision intensity codes are unknown because no ground-truth revision exists ($r_{out}$). As a proxy, we fix $input_{frac} = \frac{|\bm{r_{in}} \cap \bm{S}|}{|\bm{r_{in}}|}$, which instructs the model to remove words from the input proportional to its lexical overlap with $\bm{S}$. Then, we vary $source_{frac}$ and over-generate 10 revision candidates with different levels of copy-paste from $\bm{S}$ and re-rank each candidate to select a final revision. In this way, the codes are useful both as a control mechanism and as a prompt for diverse generation. We experiment with two different scoring functions for re-ranking, which are discussed below as part of the experimental setup. 



\paragraph{Implementation Details.} The \textbf{reviser} is trained from \texttt{bart-base} on the subset of references sentences classified as supported (417k, from \S \ref{sec:alignment}), and then used to over-generate revisions for the 595k \emph{un}supported sentences. The top scoring revision replaces the original sentence in the training data.





\section{Experimental Setup} \label{sec:experiments}

We design experiments around our central hypothesis: \emph{for a setting (long form hospital-course summarization), in which high-quality reference data is limited, reference revision is the best data-centric intervention to improve model faithfulness.} As such, we restrict the set of comparison methods to model-agnostic methods which explicitly address data quality. Based on our thorough literature review, we consider two classes of baselines: those which \textbf{filter} low quality data \citep{kang-hashimoto-2020-improved, narayan2021planning, matsumaru-etal-2020-improving, nan2021entity,goyal2021annotating}, and those which \textbf{control} for it \citep{filippova-2020-controlled}.


\paragraph{Reference Revision Strategies.} We experiment with two different functions to re-score over-generated candidate revisions: \textbf{Less Abstractive} selects the one with the highest BERTScore precision, while \textbf{More Abstractive} adds a penalty, based on the extractive fragment density \citep{grusky2018newsroom}, to encourage more abstraction. We also consider a baseline revision approach: \textbf{Fully Extractive}, which replaces each unsupported reference sentence with the source sentence with the highest BERTScore overlap. Even though our dataset is highly abstractive, this is necessary to justify the complexity of abstractive revision.

\paragraph{Baselines.} \textbf{(1) Filtered.} We experiment with three heuristics for low quality: references where no Admission Note is available in the source documents (\textbf{No Admission}) \citep{shing2021towards}, references where a significant portion of the content is unsupported by the source notes ($<0.75$ token coverage or entity hallucination rate\footnote{\citet{nan2021entity, narayan2021planning} rely on entities.} of $>10\%$) (\textbf{Unsupported}), and \textbf{Halluc. Ents}, which masks the training loss over spans of hallucinated reference entities. \textbf{Halluc. Ents} is inspired by \citet{goyal2021annotating} who use a factuality model to ignore negatively entailed dependency arcs\footnote{They use the dependency-arc entailment (DAE) model \citep{goyal-durrett-2020-evaluating} to identify inconsistent spans. Without such a model for clinical text, we use unsupported entities.}. Given the poor performance of other filtering strategies, we did not implement entailment-based filtering \citep{matsumaru-etal-2020-improving}. We also implement \textbf{Loss Truncation} \citep{kang-hashimoto-2020-improved}, which involves training for a fixed number of steps on the full training data before skipping high log loss examples. We grid-searched for the optimal number of warmup steps (2k) and the fraction of examples to drop (0.6) during truncation. \textbf{(2) Control Hallucination.} We implement the method in \citet{filippova-2020-controlled}: group training data into quality buckets based on token coverage, and control for hallucinations with encoder-prefixed style codes.



\begin{table*}[t!]
\small
\centering
\begin{tabular}{c|cc|cccccc}
& \textbf{\makecell{Reference \\ Version}} & \textbf{\makecell{Quality \\ Strategy}} & \textbf{\makecell{Hallucination\\Rate (HR) $\downarrow$}} & \multicolumn{3}{c}{\textbf{\makecell{BERTScore\\P / R / F1 (BS) $\uparrow$}}} & \textbf{\makecell{Entail. \\$\uparrow$}} & \textbf{\makecell{Faithful-Adjusted \\ Recall (FaR) $\uparrow$}} \\ \hline 
\hline
\parbox[t]{1ex}{\multirow{9}{*}{\rotatebox[origin=c]{90}{\sc{\textbf{{Longformer}}}}}} & 
\multirow{2}{*}{\textbf{Original}} & N/A & \textbf{36.8} & \textbf{82.3} & 69.5 & 75.2 & \textbf{48.4} & \textbf{48.2} \\ 
& & Control Halluc. & 36.5 & 83.3 & 70.2 & 76.0 & 51.5 & 49.0 \\ \cline{2-9}. 
& \multirow{4}{*}{\textbf{\makecell{Filtered \\ (Baselines)}}} & No Admission & 20.1 & 87.8 & 70.4 & 78.0 & 61.6 & 41.2 \\ 
& & Unsupported & \textbf{18.4} & 87.6 & 70.8 & 78.1 & \textbf{61.6} & \textbf{46.9} \\ 
& & Loss Truncation & 36.3 & 83.1 & 69.9 & 75.8 & \textbf{51.7} & 47.4 \\  
& & Halluc. Ents & 33.8 & 83.5 & 69.8 & 75.9 & 55.0 & 47.7 \\ \cline{2-9}. 
& \multirow{3}{*}{\textbf{\makecell{Revised\\(Ours)}}} & Fully Extractive & \textbf{5.4} & 94.5 & 73.2 & 82.3 & 78.6 & \textbf{52.4} \\ 
& & Less Abstractive & \textbf{3.8} & \textbf{94.6} & 73.1 & 82.3 & \textbf{83.7} & \textbf{54.0} \\ 
& & More Abstractive & \textbf{5.6} & 92.1 & 73.0 & 81.3 & 76.3 & \textbf{57.1} \\ \hline 
\end{tabular}
\caption{Summarization quality metrics across reference quality mitigation strategies (original, filtered, control, revised). The Longformer Encoder-Decoder (LED) model is used for fine-tuning. \textbf{Numbers} discussed below.}
\label{tab:results}
\end{table*}

\paragraph{Training Details.} We fine-tune downstream summarization models from BART \citep{lewis2019bart} and the encoder-decoder Longformer \citep{beltagy2020longformer}. We train all models for 10,000 steps or until convergence on the validation set. Some methods use revised or filtered training data yet all use the same 1,195 validation examples and evaluation test set (1,190). Please refer to Appendix \ref{app:sum-train-detail} for hyperparameters and more training details.


\paragraph{Metrics.} To measure source faithfulness, we compute the entity \textbf{hallucination rate} (HR) using the soft matching heuristic described in \S \ref{app:entity-merging}, \textbf{BERTScore} (BS) using in-domain weights from Clinical BERT \citep{alsentzer2019publicly}, and the fraction of summary sentences predicted as \textbf{entailed} by SciFive \citep{phan2021scifive} fine-tuned on MedNLI \citep{romanov2018lessons}\footnote{MedNLI is a clinician-annotated entailment dataset whose premise sentences come from MIMIC-III. SciFive is a biomedical T5 model that achieves SOTA performance on MedNLI.}. To capture entity coverage, we record \textbf{faithful-adjusted recall} (FaR): the fraction of non-hallucinated reference entities included in the output~\citep{shing2021towards}. 



\section{Results} \label{sec:sum-results}

Please refer to Appendix for basic statistics on the revised training datasets (\S \ref{app:revisedtraining-data}), BART summarization results (\S \ref{app:bart-results}) and an example of over-generated revisions (by varying the $source_{frac}$ code from \S \ref{sec:reviser}), which enables us to optimize the abstractive-faithful tradeoff during revision re-ranking (\S \ref{app:qualitative}).


\paragraph{Impact of Revisions on Summarization.}
Table \ref{tab:results} confirms that filtering improves faithfulness (\textbf{Filtered - Unsupported} lowers the HR from 36.8 to 18.4 and improves entailment from 48.4 to 61.6), yet degrades coverage (48.2 vs 46.9 FaR). Masking hallucinated entities from log loss training (\textbf{Filtered - Halluc. Ents}) only slightly improves faithfulness, which underscores the difficulty in assigning token-level credit to pervasively noisy references. \textbf{Loss Truncation} leads to worse performance except on entailment (48.4 vs 51.7), which can likely be attributed to: 1) learning from fewer examples (from truncation); 2) hallucination patterns learned from the full data warmup are not unlearned during truncation; and 3) log loss might not capture faithfulness as well as more direct measures, i.e., entity overlap (used by \textbf{Halluc. Ents}).

In comparison, all revision approaches yield dramatic improvements in faithfulness (e.g., for \textbf{Less Abstractive}, 3.8/94.6.3/83.7 vs 36.8/82.3/48.4). These precision-oriented gains do not come at the expense of coverage (as defined by FaR), which actually jumps from 48.2 to 54.0. Abstractive revision (\textbf{Less} and \textbf{More}) outperforms \textbf{Fully Extractive} on coverage (e.g., 54.0/57.1 vs 52.4). Surprisingly, despite Fully Extractive revision being perfectly faithful, Less Abstractive revision leads to more faithful models (e.g., 3.8/83.7 vs 5.4/78.6 for HR and entailment, respectively), which suggests the need to re-write, not replace, unsupported content. Out of the revised results, More Abstractive has the best coverage (56.3/57.1) while being competitive on faithfulness (5.6 vs 5.4/3.8 HR).




\begin{table}[ht]
\small
\centering
\setlength\tabcolsep{3.5pt}
\begin{tabular}{lc|cccc}
\textbf{\makecell{Reference \\ Version}} & \textbf{\makecell{Quality \\ Strategy}} & \textbf{Con.} & \textbf{Rel.} & \textbf{Fl.} & \textbf{Coh.} \\ \hline
\textbf{Original} & N/A & 1.5 & 2.2 & 3.9 & \textbf{3.5} \\
\textbf{Filtered} & Unsupported & 3.2 & 2.8 & 3.3 & 3.3 \\
\textbf{Revised} & More Abstractive & \textbf{3.5} & \textbf{3.0} & \textbf{4.0} & 2.7 \\
\end{tabular}
\caption{Average rating (1-5) assigned by a domain expert (clinician) according to Consistency (Con.), Relevance (Rel.), Fluency (Fl.), and Coherence (Coh.).}
\label{tab:results-human}
\vskip -0.1in
\end{table}

\paragraph{Human Evaluation.} We rely on the protocol from \citet{fabbri2021summeval} to procure expert judgments on \emph{Consistency}, \emph{Relevance}, \emph{Fluency}, and \emph{Coherence}. An experienced clinician was shown the source notes for 10 examples and 3 randomly shuffled Longformer outputs side-by-side (10 x 3), and asked to rank each summary on a 1-5 Likert Scale. Please refer to Appendix (\S \ref{app:human-eval}) for more detail on the protocol. We include the most faithful baseline according to automatic metrics: \textbf{Filtered - Unsupported}, as well as \textbf{Original} and \textbf{Revised}.

Table \ref{tab:results-human} shows that the model trained on Abstractively revised references produces the most Consistent, Relevant, \emph{and} Fluent summaries. Training on Filtered references improves consistency and relevance but degrades fluency given the relative paucity of training data (a similar finding for data-to-text generation is noted by \citet{filippova-2020-controlled}). Assessed coherence is lower for models trained on Abstractively Revised data, yet we note that for 3/10 summaries, it held the highest coherence rating. Additionally, there was large variance in coherence ratings within each system (1.34 standard deviation), as each had at least one summary rated as 1 and at least one as 5. Consistency had the lowest (1.04). \citet{adams-etal-2021-whats} argue that discourse is hard to evaluate because ``clinical summaries naturally exhibit frequent, abrupt topic shifts''.




\begin{figure*}[t]
\centering
\includegraphics[width=\linewidth]{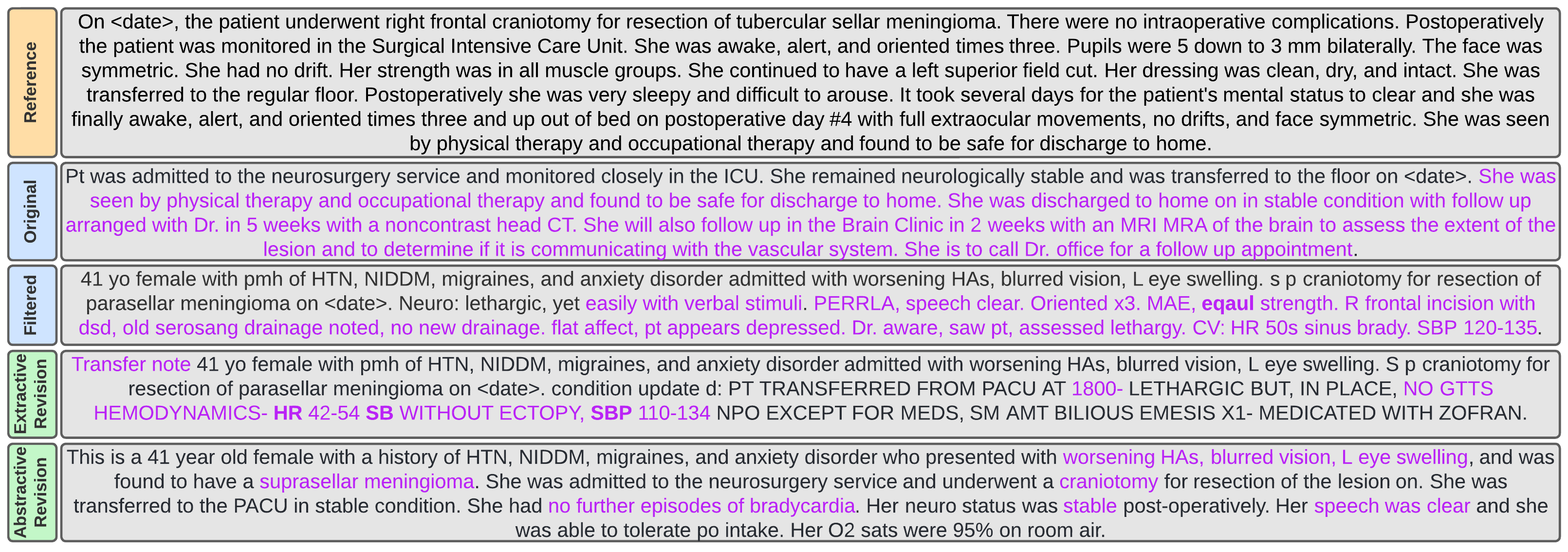}
\caption{An example \textcolor{orange}{Reference}, along with outputs from models trained on un-revised references (\textcolor{blue}{Original} and \textcolor{blue}{Filtered}), as well as \textcolor{Green}{Extractive} and \textcolor{Green}{Abstractive} revisions. Times and ages have been modified to ensure anonymity. Source notes are not shown for space reasons. \textbf{\textcolor{Amethyst}{Purple Snippets}} represent text mentioned in our analysis below.}
\label{fig:sum-output-example}
\end{figure*}

\paragraph{Qualitative Analysis.} Figure \ref{fig:sum-output-example} shows an example from the human annotation test set for which we requested additional qualitative feedback. ``Original'' was inconsistent because it fabricates the last 4/6 sentences, operating almost as an unconditional LM. ``Filtered'' was less relevant than ``Abstractive'' because the second half of Filtered (starting with ``PERRLA'') contains non-essential status information. ``Filtered'' fluency was discounted for a misspelling (``eqaul'' rather than ``equal'') and a missing word (``easily \emph{roused} with verbal stimuli'').


The annotator noted high fluency for Abstractive Revision (5) because the summary mimics the narrative style of the original references, rather than a sequence of choppy lists as in Filtered and Extractive summaries. Consistency is high (5) because the facts presented can be directly inferred. Coherence and consistency scored 5 because it follows a conventional arc from presenting illness (headache, vision, eye swelling), diagnosis (meningioma) and treatment (craniotomy), and ends with the patient's status (stable - no bradycardia, clear speech).



We also include ``Extractive'' to further emphasize the need to train on abstractive revisions. The ``Extractive'' summary does not match the target style of the Brief Hospital Course and includes many undesirable artefacts from source notes: note type metadata (``Transfer note''), long lists of non-essential data points (HR, SB, SBP), and unclear section boundaries (``1900- Lethargic...'').

To better understand slightly lower coherence results for ``Abstractive Revision'', we include an extra example in the Appendix (\S \ref{app:qualitative}) for which coherence was rated as a 3. We hypothesize that it stems from a sentence-level revision strategy which does not consider the position in the summary. As such, heavily revised references can deviate slightly from a conventional structure for the hospital course.

\begin{table}[ht]
\small
\setlength\tabcolsep{1.5pt}
\centering
\begin{tabular}{l|cc}
\textbf{\makecell{Reviser Training Objective \\ (Each Ablation is Separate)}} & \textbf{\makecell{BERTScore\\P / R / F1 (BS) $\uparrow$}} & \textbf{\makecell{Entail. \\$\uparrow$}} \\ \hline
\textbf{Full} (last row of Table \ref{tab:results}) & \textbf{92.1} / \textbf{73.0} / \textbf{81.3} & \textbf{76.3} \\ \hline
\textbf{\emph{w/o}} ReDRESS hallucinations & 90.8 / 72.4 / 80.4 & 72.1 \\
\textbf{\emph{w/o}} Random \textit{other} alignments & 90.9 / 72.8 / 80.7 & 73.8 \\
\textbf{\emph{w/o}} All Negatives (no contrast) & 90.9 / 72.8 / 80.7 & 72.4 \\
\textbf{\emph{w/o}} Revision Codes & 87.6 / 71.7 / 78.7 & 62.5 \\
\end{tabular} 
\caption{Separately removing key components of the reviser training objective (from Equation \ref{eq:revision}) hurts the \emph{downstream} performance of Longformer summaries.}
\label{tab:results-ablation}
\end{table}

\paragraph{Ablation Analysis.} Which parts of the revision pipeline are necessary to improve \emph{downstream} summary performance? Separately, we remove ReDRESS hallucinations (top sequences from Figure \ref{fig:revise-train-example}), randomly sampled \emph{other} alignments (bottom sequences), and all negative examples (right side). We also train a model without control codes for revision intensity (\emph{w/o} Revision Codes) and, in turn, at inference time, generate a single revision rather than vary the codes to over-generate-then-rescore. Table \ref{tab:results-ablation} reveals that both sources of synthetic revision training data contribute to downstream performance (BS F1/Entailment of 80.4/72.1 and 80.7/73.8, individually, vs 81.3/76.3 combined). This 
aligns with \citet{cao2021cliff} who also demonstrate "the importance of covering diverse types of errors in negative samples". Eliminating unlikelihood training of synthetic negatives reduces summary BS F1/Entailment from 81.3/76.3 to 80.7/72.4. Removing the ability to over-generate-then-rescore diverse revision candidates by varying style codes (\emph{w/o} Revision Codes) has the largest impact on downstream performance (78.7/62.5). A model trained without codes tends to insufficiently edit highly unsupported sentences.

\begin{figure}[t]
\includegraphics[width=\linewidth]{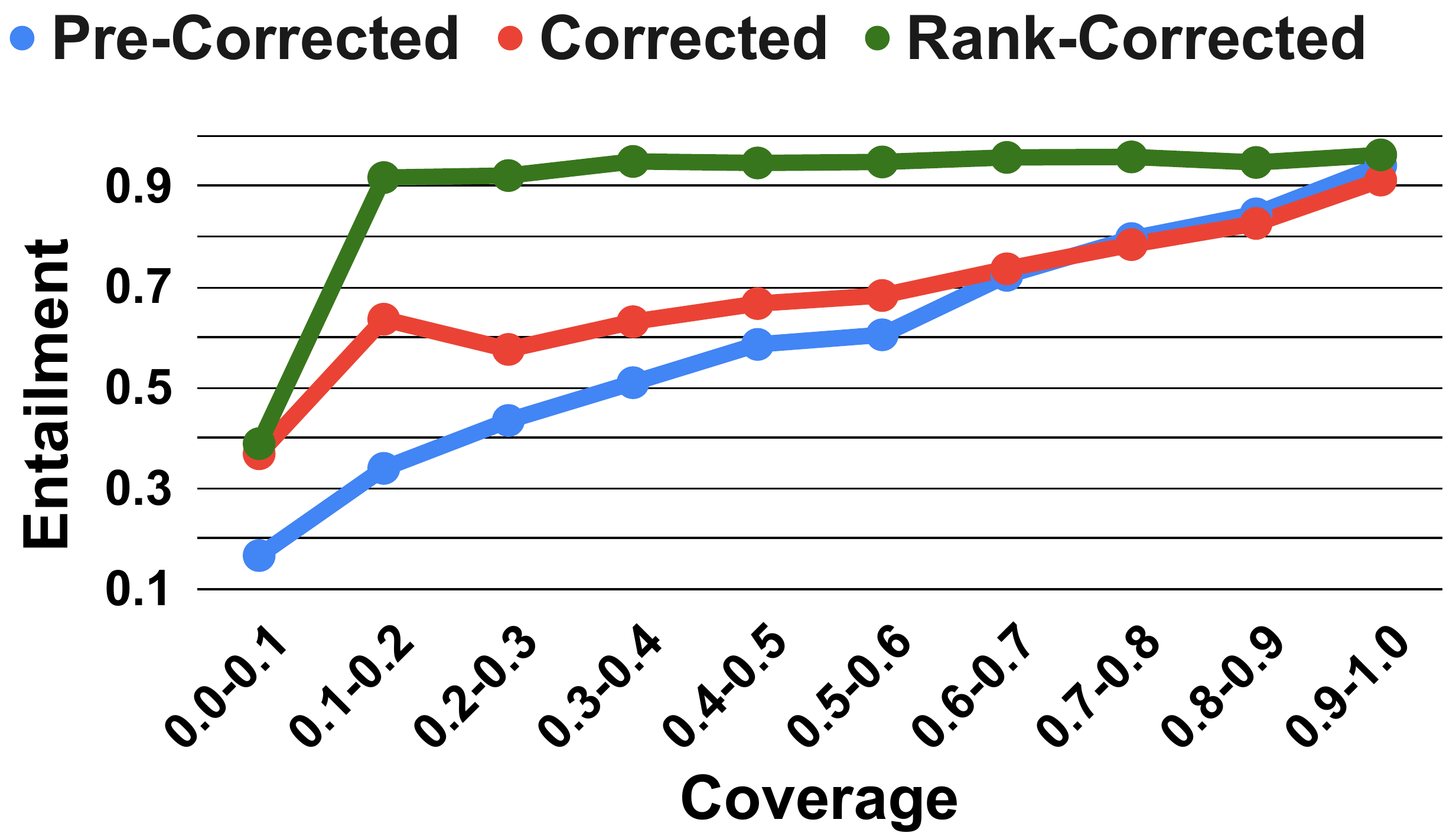}
\caption{Faithfulness before and after correcting summaries with the Reviser, controlling for extractiveness.}
\label{fig:effective-faithfulness}
\centering
\vskip -0.12in
\end{figure}

\begin{table*}[t]
\small
\centering
\begin{tabular}{l|cccccc}
 \textbf{\makecell{Pretrain\\Weights}} & \textbf{\makecell{Hallucination\\Rate (HR) $\downarrow$}} & \multicolumn{3}{c}{\textbf{\makecell{BERTScore\\P / R / F1 (BS) $\uparrow$}}} & \textbf{\makecell{Entail. \\$\uparrow$}} & \textbf{\makecell{Faithful-Adjusted \\ Recall (FaR) $\uparrow$}} \\ \hline 
\textbf{\texttt{bart-base}} & 25.6 & 85.6 & 70.7 & 77.3 & \textbf{56.0} & 44.4 \\ \hline 
\textbf{ReDRESS} & \textbf{22.3} & 86.0 & 70.7 & 77.5 & \textbf{55.2} & \textbf{44.7} \\ 
\textbf{\emph{w/o} Entity Swap} & \textbf{26.9} & 85.7 & 70.3 & 77.1 & 54.4 & \textbf{42.1} \\ \hline 
\textbf{Reviser} & 23.0 & 86.0 & 71.0 & 77.7 & 56.7 & 47.9 \\ 
\end{tabular}
\caption{Assessing the usefulness of ReDRESS \& reviser models for pre-training. We separately fine-tune models from \texttt{bart-base}, ReDRESS, and reviser checkpoints on \emph{Filtered - No Admission} data. \emph{w/o} Entity Swap restricts noise to span-deletion, similar to the optimal configuration in the original BART paper. \textbf{Numbers} discussed below.}
\label{tab:results-pretrain}
\vskip -0.12in
\end{table*}

\paragraph{Reviser as a Post-Hoc Editor.} As a natural extension, we experiment with using the revision model to correct system outputs. In particular, we take summaries from the model trained on \textbf{Original} references, and then separately feed each predicted sentence and aligned context to the reviser for generation. It would be easy to increase faithfulness by copying the context verbatim. Yet, we are more interested to see if the reviser can increases \emph{effective} faithfulness \citep{ladhak2021faithful}, i.e., controlling for extractiveness. For each \emph{predicted} sentence, we over-generate revision candidates with different targeted levels of extraction (by varying the $source_{frac}$ code defined in Section \ref{sec:revise-pipeline}. Then, we bin sentences by extractiveness (coverage) and record faithfulness (fraction of sentences predicted as entailed) within each bin. We include separate plots for \textbf{Corrected}, which includes each over-generated candidate, and \textbf{Rank Corrected}, which selects the top candidate by entailment prediction (breaking ties according to the most abstractive). Figure \ref{fig:effective-faithfulness} demonstrates that reviser-corrected summaries are much more effectively faithful, although, naturally, the gap shrinks as all outputs become highly extractive and mostly entailed. The ability to re-rank and select from diverse candidates makes a huge difference, as evidenced by the large gap between \textbf{Corrected} and \textbf{Rank Corrected}.

\paragraph{ReDRESS/Revision as Pre-Training Objectives.} Beyond data cleaning, do our proposed methods to generate hallucinations (ReDRESS), and then edit them out (reviser), hold intrinsic value for the task as pre-training objectives? One can argue that both models are trained on faithfulness objectives: based on context, ReDRESS learns to add/remove/integrate entities, and the reviser to edit out synthetic hallucinations. Table \ref{tab:results-pretrain} shows that fine-tuning from ReDRESS and reviser--both trained from checkpoints of \texttt{bart-base}--improves all evaluation metrics vis-a-vis \texttt{bart-base} (except slight entailment decrease from ReDRESS (56.0 to 55.2)). \emph{w/o} Entity Swapping is a denoising baseline in which corruptions are limited to span deletion and word re-ordering. On its own, then, incorporating entity swaps into pre-training (the full version of the ReDRESS framework), causes HR to drop (26.9 to 22.3) and NSP/FaR to rise (81.1/42.1 to 83.3/44.7).

\section{Limitations} \label{sec:limitations}

Our sentence-level revision strategy does not consider the impact of changing one sentence on the reference as a whole, which can impact coherence. This can be addressed in future work by exploring summary-level revision (especially for corpora with short summaries) or incorporating coherence into revision re-scoring and/or contrast set creation. Given the complexity and expertise required for the task, the length of both source notes and summaries, and a mandatory license for data access, only 30 human ratings were procured from a single annotator. The assessment still took four full days.



\section{Conclusion}

We propose a new approach to mitigating the downstream impact of noisy references on summarization models. We learn a model to improve the coverage of existing references from a large clinical corpus of EHR notes, and re-train models on revised references. Results show that reference revision is a more effective intervention for improving faithfulness than quality filtering for this task.  



\section{Ethical Considerations}

\paragraph{Deidentification.} Our summarization corpus is extracted from a publicly avilable database of real-world, de-identified clinical records: MIMIC-III v1.4~\citep{johnson2016mimic}. Even though it is HIPAA-compliant, we make sure no Protected Health Information (PHI) is shared with the public.

\paragraph{Intended Use \& Failure Modes.} The goal of this paper is to make progress toward automatic summarization of a patient's hospital admission. Deploy such a system in a real-world clinical setting has its own set of ethical and procedural concerns. Robustness, Fairness, and Trust is vital to any NLP system, especially one deployed in a hospital setting where lives are at risk. As with many NLP datasets, our MIMIC-III dataset likely contains biases, which may be perpetuated by its use. It is important to analyze the underlying population to identify demographic, social, and economic discrepancies vis-a-vis the broader population of interest. Model-generated errors could be harmful to patient safety and even negatively affect outcomes. There are lessons to be learned from existing clinical decision support tools \citep{pivovarov2016can, chen2020ethical}. Furthermore, there is a moral hazard from deploying clinical systems, in which clinicians start to over-rely on a system at the expense of their own judgment \citep{goddard2012automation}. EHRs are also living systems and deploying a summarization system within it necessitates evolving with the EHR and the underlying population.

\section*{Acknowledgments}

We thank the reviewers for the insightful, actionable insights, as well as Michael Elhadad and Alex Fabbri for their feedback on earlier drafts.


\bibliographystyle{acl_natbib}
\bibliography{anthology,my_full}

\appendix

\section{Merging Entities} \label{app:entity-merging}

For each pair of entity mentions in the source text and reference, we compute a code overlap score. Let entity $e_x$ have codes $\bm{c_x}$ and tokens $\bm{t_x}$ and $e_y$ have codes $\bm{c_y}$ and tokens $\bm{t_y}$, the pairwise code overlap score is:

$$
CodeOverlap(e_x, e_y) = \frac{|\bm{c_x} \cap \bm{c_y}|}{|\bm{c_x}| + |\bm{c_y}|}
$$

Then, we compute embedding cosine similarity between mention tokens with BioWordVec \citep{biowordvec}, filtering out stopwords and punctuation. Let

$$
EmbedOverlap(e_x, e_y) = cosine(E(t_x), E(t_y))
$$

Finally, we compute the TF-IDF overlap ($TF\_IDF(t_x, t_y)$) to compensate for occasional noise in embedding space. We define the aggregate score as the average of embed, code, and TF-IDF overlaps. For entity types with no available codes (treatments, procedures, and tests), we only examine mention overlap. Based on validation on a manually labeled held-out set, we classify $e_x$ and $e_y$ as synonyms iff any of the following thresholds are met: $CodeOverlap(e_x, e_y) \geq 0.4$, $EmbedOverlap(e_x, e_y) \geq 0.75$, or $AggOverlap(e_x, e_y) \geq 0.4$.

\subsection{Analyzing Unsupported References} \label{app:hallucination}

\paragraph{Identifying Correlates}

\begin{figure}[ht]
\includegraphics[width=\linewidth]{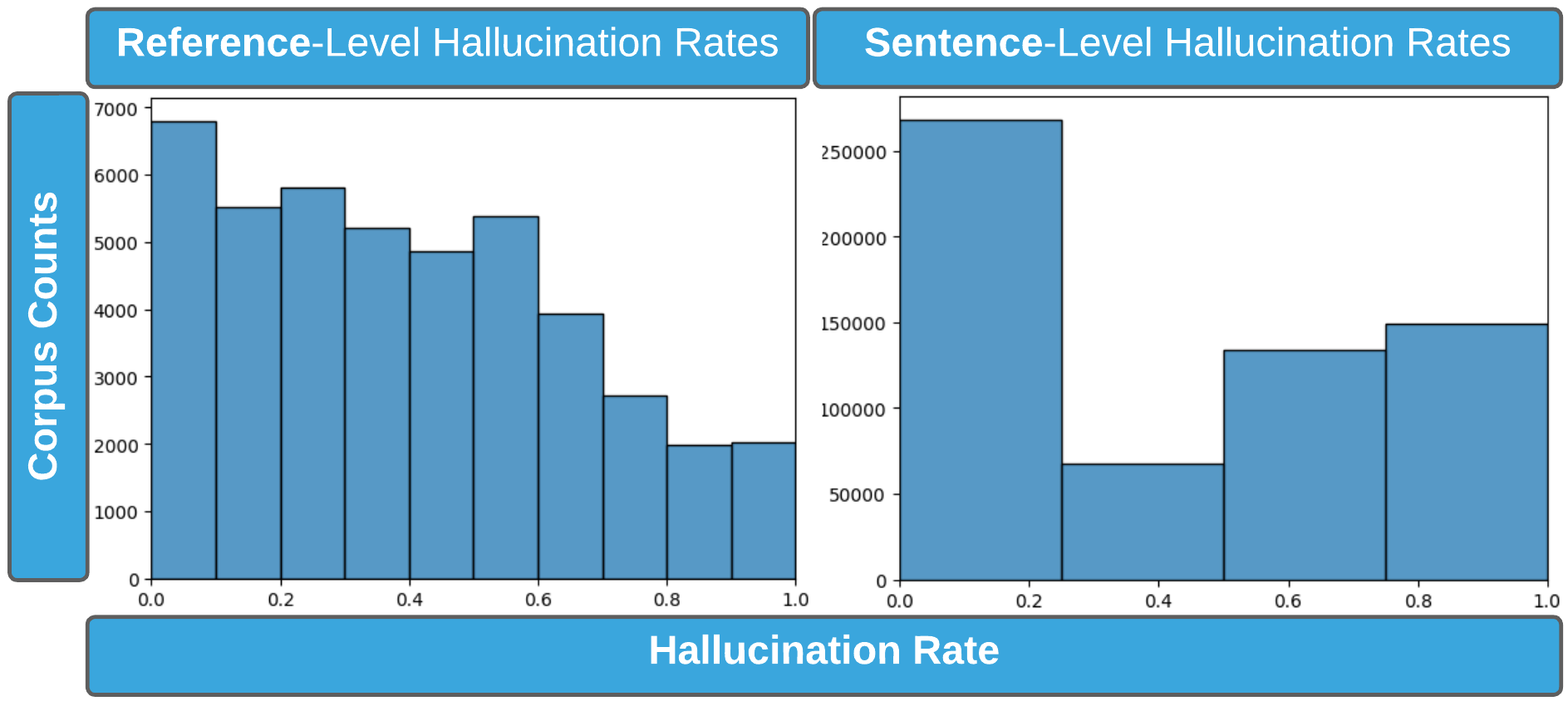}
\caption{The distribution of reference-level (left) and reference \textit{sentence}-level (right) hallucination rates--the fraction of entity mentions not present in source text.}
\label{fig:halluc-dist}
\centering
\end{figure}

We collect a wide-range of visit-level statistics and compute the Pearson correlation coefficient with respect to the hallucination rate, defined as the fraction of entities in the reference text not present in the available source documentation. Unsurprisingly, lexical overlap (unigram coverage) is highly correlated (88\%) with the hallucination rate. Examples with well-supported references contain more source notes and distinct note types. The MIMIC-III notes do not cover time a patient spent outside the ICU. Interestingly enough, however, the number of days spent outside the ICU has zero direct correlation with the hallucination rate.

\paragraph{Distribution of Hallucinations.} We examine the example and sentence-level distribution of hallucinations before devising a revision strategy. Figure \ref{fig:halluc-dist} reveals a degree of uniformity in the example-level hallucination rate and, to a lesser extent, sentence-level\footnote{Sentence-level hallucination rates are shown only for multi-entity sentences to get a sense of the non-binary distribution. $14\%$ of reference sentences have no entities, and 30 \% have one. Single entity sentences have a hallucination rate of 41\%.}. The figure indicates that the faithfulness issue is not concentrated to just a few very low coverage examples, or sentences. As such, example-level quality filtering is noisy since there is no clear coverage boundary and relatively few references contain zero entity hallucinations ($<2k$). These two basic plots inform two key design choices: \textbf{(1.)} to address quality at the sentence-level rather than the reference; \textbf{(2.)} to enforce diversity of faithfulness in synthetic hallucinations.

\begin{figure*}[t]
\centering
\includegraphics[width=0.75 \linewidth]{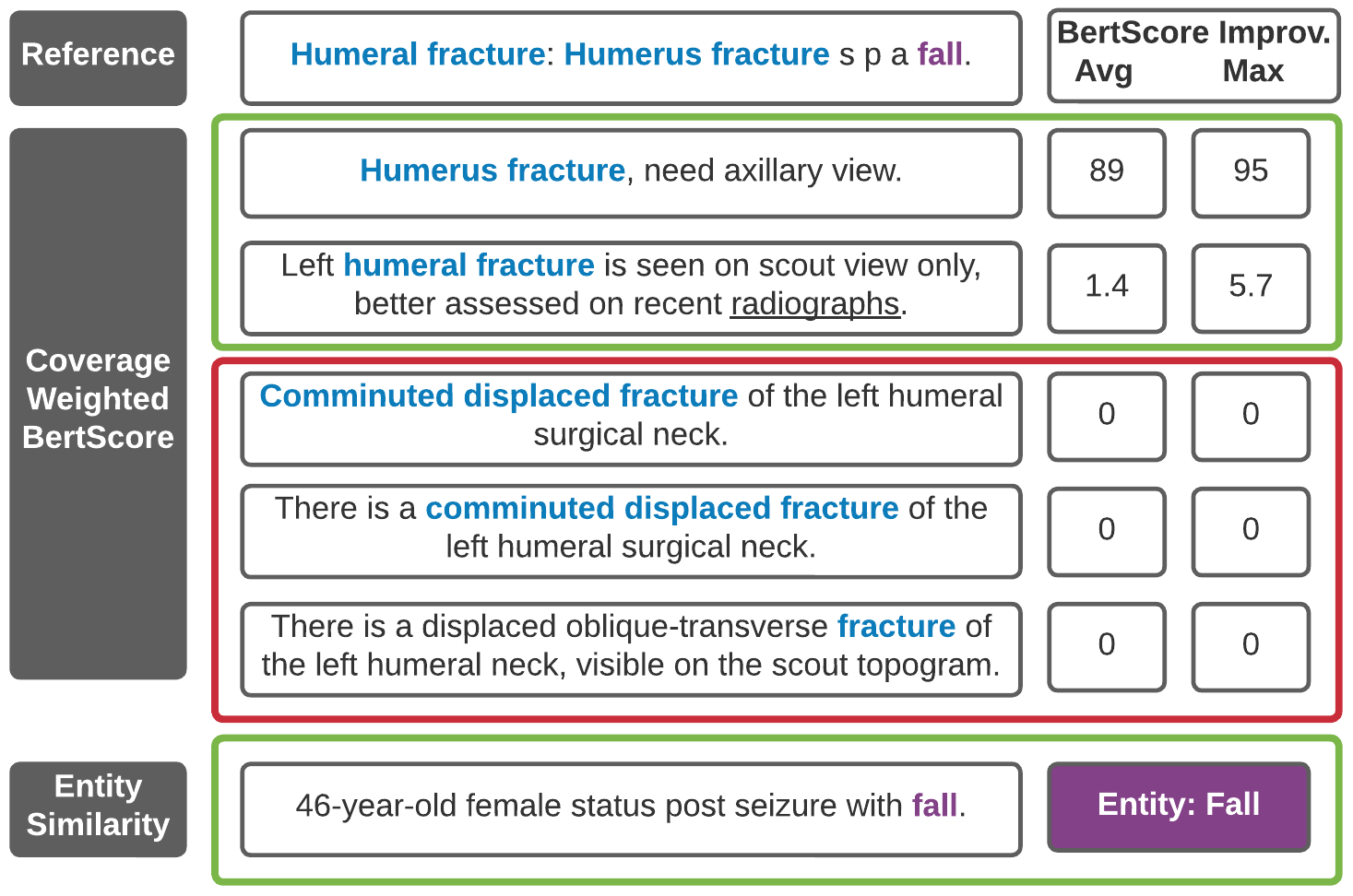}
\caption{Source-Reference Alignment. 5 source sentences are greedily extracted with a coverage-weighted BERTScore heuristic. Non-influential sentences are \textcolor{red}{discarded} (low BERTScore improvement). Finally, in the case of missing clinical concepts (\textbf{fall}), we find the source sentence whose contextualized representation of the entity span (\textbf{fall}) is closest to the reference usage (the last word of the top reference row). }
\label{fig:alignment}
\end{figure*}

\section{Alignment Algorithm} \label{app:alignment}

Figure \ref{fig:alignment} provides an example alignment with improvement filtering and an extra extraction step to ensure full entity coverage.

\subsection{Notation} \label{sec:notation}

Let $\braket{(\bm{S_1}, \bm{R_1}), ..., (\bm{S_N}, \bm{R_N})}$ represent the corpus, consisting of source-reference pairs for $N$ unique patient ICU admissions. Let $\bm{S_n} = \braket{\bm{s_1^{n}}, ..., \bm{s_{|S_n|}^{n}}}$ represent the sentences extracted from the source input for the $n^{th}$ example and, likewise, $\bm{R_n} = \braket{\bm{r_1^n}, ..., \bm{r_{|R_n|}^n}}$ the reference sentences. Similarly, $\bm{s_i^{n}} = \braket{x_1, ..., x_{|s_i^n|}}$ is the tokenized sequence for the $i^{th}$ source sentence from the $n^{th}$ example, and $\bm{r_j^{n}} = \braket{\hat{x}_1, ..., \hat{x}_{|r_j^{n}|}}$ the tokenization of the $j^{th}$ reference sentence.

Given very long inputs, we link each reference sentence $\bm{r_{j}^{n}}$ ($n \in N$, $j \in \bm{R_{n}}$), to a small subset ($\leq 5$) of source sentences corresponding to the same example. Due to the abstractiveness of the data (acronym usage, shorthand, etc.), as well as redundancy from copy-and-paste \citep{hirschtick2006copy}, we align sentences using a new approach which combines BERTScore and subword-level coverage, rather than the conventional approach of lexical overlap with ROUGE \citep{lebanoff2019analyzing, liu2019hierarchical}. Given a candidate alignment pair: a reference sentence $\bm{r_j^n}$ with $K$ tokens and a source sentence $\bm{s_i^n} \in \bm{S_n}$ with $L$ tokens, for each reference token $\hat{x}_{k}$, we find its closest match in $\bm{s_i^n}$:

$$
align(\hat{x}_k, \bm{s_i^n}) = \underset{1 \leq \ell \leq L}{\max}cos(h(\hat{x}_k), h(x_{\ell}))
$$

where $h(x)$ represents the contextualized BERT embedding\footnote{The mean-pool of the last four layers of ClinicalBERT.}. Based on these greedy alignments, we extract sentences for $T$ steps. At $t=0$, we initialize an importance vector $\bm{w}$ of length $K$, to all ones. Then, at each timestep, we select $\bm{s^*} \in \bm{S_n}$ which maximizes the importance-weighted BERTScore:

$$
\bm{s^*} = \underset{\bm{s^*} \in \bm{S_n} }{\mathrm{argmax}}({\frac{\sum_{k=1}^{K}{w_{tk} align(\hat{x}_k, \bm{s^*})}} {\sum_{k=1}^{K}{w_{tk}}})}
$$

After each timestep, we keep track of the best alignment score for each reference token via the importance vector. In particular, we update each token's importance by the inverse of its best coverage score: $w_{t+1,1} = min(w_{t1}, 1 - align(\hat{x}_1, \bm{s^*}))$ (formula shown for first element). Similarly to Maximal Marginal Relevance (MMR) \citep{carbonell1998use}, the importance vector de-prioritizes well-covered tokens for future extractions. We also remove $\bm{s^*}$ from $\bm{S_n}$ to ensure it is only extracted once. After retrieving the top $K$ using this procedure, we only use sentences for the final alignment set for which the average coverage improvement $\geq 0.01$ or the max $\geq 0.05$, where improvement is defined as the reference token-level increase in coverage of the latest extraction over the previous max from prior extractions: $max(0, \bm{w^t} - \braket{align(\hat{x}_1, s^*), ..., align(\hat{x}_K, s^*)})$.

Infrequently, the medical concepts extracted from the aligned sentences do not cover all the concepts in the reference sentence. In this case, for each missing concept, we filter for the subset of source sentences containing that concept, and add the sentence with the highest pairwise similarity of contextualized concept embeddings--the mean of hidden states corresponding to the entity span.

\begin{table*}[t]
\centering
\small
\begin{tabular}{lccccc}
\hline
\textbf{Model} & \textbf{\makecell{Hallucination \\ Rate (HR) $\uparrow$}} & \textbf{\makecell{BERTScore \\ F1 (BS) $\downarrow$}} & \textbf{\makecell{Coherence \\ (NSP) $\uparrow$}} & \textbf{\makecell{Diversity \\ $\uparrow$}} \\
\hline
\textbf{Swap Random} (Baseline) & 45 & 91.5 & 70.0 & 16 \\
\textbf{Swap Related} (Baseline) & 32 & 93.2 & 73.9 & 14 \\ \hline
\textbf{Span Infill + Entity Swap} (ReDRESS) & 38 & 87.0 & 72.6 & 23 \\
\textbf{\emph{w/o} Entity Swap} & 21 & 90.1 & 67.7 & 21 \\
\textbf{\emph{w/o} Add-1 Inference Trick} & 28 & 88.8 & 72.4 & 22 \\
\textbf{\emph{w/o} Entity Hiding Inference Trick} & 27 & 89.7 & 73.4 & 19 \\
\end{tabular}
\caption{Intrinsic evaluation of ReDRESS model (Span Infill + Entity Swap) on \emph{supported} reference sentences. For an upper bound, the original sentences have coherence of $75.8$. We define \textbf{diversity} as one minus the pairwise unigram coverage score \citep{grusky2018newsroom} between two synthetic hallucination samples for the same input. Greater error diversity has been shown to be useful for synthetic data generation in other summarization work \citep{cao2021cliff}. Each ablation (\emph{w/o}) is performed independently of the others.}\label{tab:perturb-results}
\end{table*}

\begin{figure*}[t]
\includegraphics[width=\linewidth]{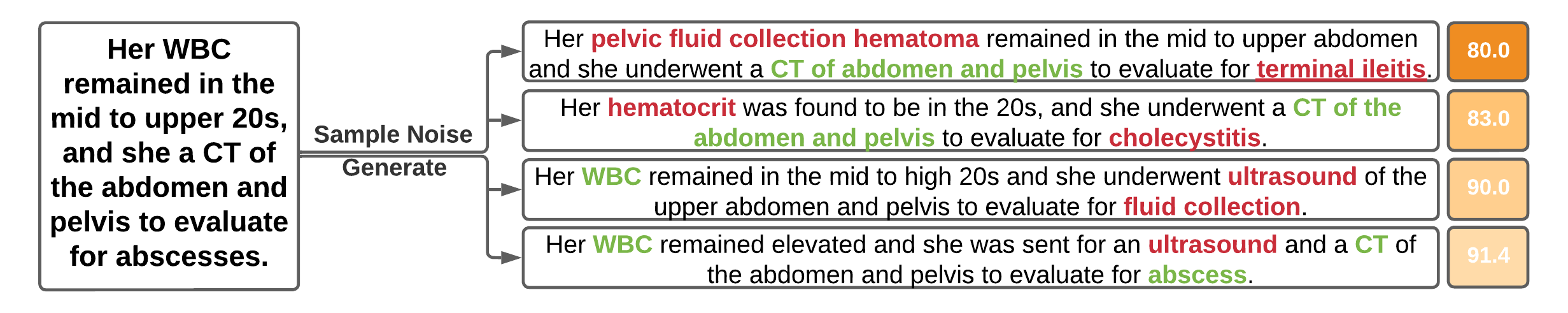}
\caption{\textbf{ReDRESS} model outputs. \textcolor{Green}{Green} represents non-hallucinated entities, \textcolor{red}{red} not present in input sentence, and \underline{\textcolor{red}{red}} also not present in any of the source notes. The orange box shows BERTScore F1 vis-a-vis original. Due to the topical nature of the distractor set, all hallucinations except \textit{terminal ileitis} exist elsewhere in the source notes.}
\label{fig:perturb-output}
\centering
\end{figure*}


\section{ReDRESS Details} \label{app:perturber-train-details}


\paragraph{Pre-Training Data.} We pre-train on a large unlabeled sentence corpus: all sentences extracted from MIMIC-III discharge summaries, excluding notes related to patients in the summary test set. To minimize EHR-related noise, we filter out sentences without any clinical concepts and those found in non-narrative sections related to structured data, demographics, and/or administration (billing codes, dates, times, signatures, lab values).

\paragraph{Intrinsic Evaluation of ReDRESS.} \textbf{ReDRESS} combines entity swapping and span-infilling--so we compare it approaches that do one or the other. For pure entity swaps: \textbf{(1.) Swap Random} randomly removes entities and replaces them with a random one of the same type from the training data. Given the long tail of rare entities, we sample the replacement entity by its empirical frequency in the corpus. \textbf{(2). Swap Related} follows an identical procedure with the exception that replacement entities are sampled from the related distractor set. \textbf{Span Fill} is a version of ReDRESS model without entity swaps (and no pre-pended distractor set). Corruption is limited to span removal and word order shuffling. Each baseline/ablation follows the same approach: over-generate five candidate hallucinations by re-sampling noise levels.

\begin{figure*}[t]
\centering
\includegraphics[width=\linewidth]{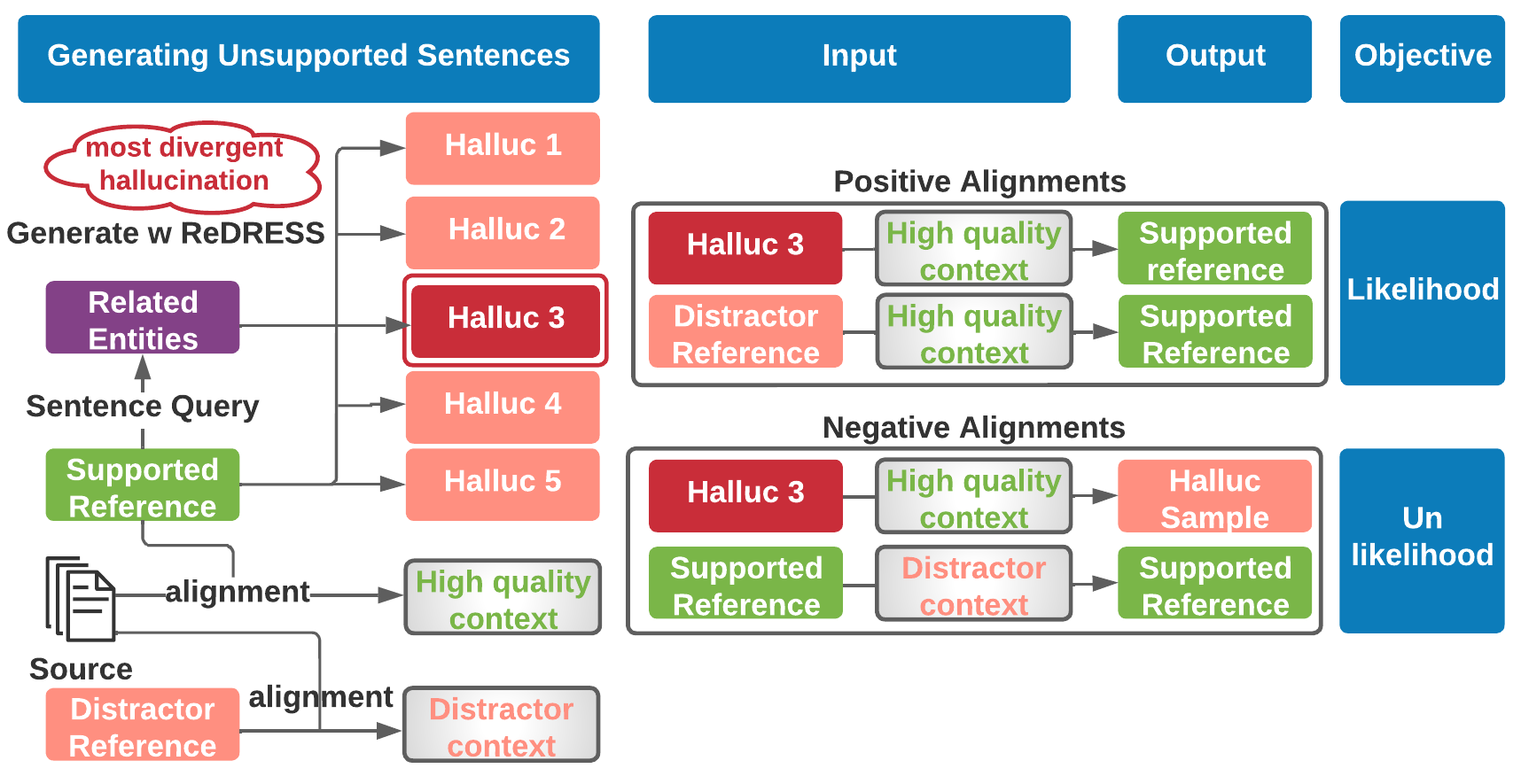}
\caption{A high-level depiction of the \textbf{reviser} training pipeline, which includes generating data from ReDRESS hallucinations and sampling within-example mis-alignments (a distractor reference and distractor context).}
\label{fig:reviser-architecture}
\end{figure*}

Based on Table \ref{tab:perturb-results}, the unified ReDRESS model--\textbf{Span Infill + Entity Swap}--achieves diverse hallucinations while maintaining topical consistency. Interestingly, the prepended distractor set greatly improves the coherence of the outputs (\textbf{Span Fill} vs. \textbf{Span Fill + Ent Swap}) because the distractor set is topically consistent. Entity swaps alone produce incoherent sentences. The corpus-level hallucination rate is 40\% which we are nearly able to achieve (38\%) with ReDRESS while maintaining topicality. The first ablation (\emph{w/o} Entity Swap) removes entity swapping entirely from training and inference. This dramatically reduces the rate of hallucinations (38 to 21) and reduces coherence dramatically. Coherence is reduced because the topical entities from the distractor set help to keep the generation on track. More specifically, the pre-pended set allows us to not have to include source context, as in \citet{cao2021cliff}, to avoid excessive semantic drift from mask-and-fill generation. We also can see that the \emph{Add-1} inference trick is working as expected. Removing it (\emph{w/o Add-1} Inference Trick) leads to 10\% lower hallucination rate (38 vs 28) without compromising coherence. In other words, we instruct the model to implement an additional entity swap and it appears to be doing so. Figure \ref{fig:perturb-output} demonstrates the diversity from the standpoint of metrics (BERTScore) and meaning.


\section{Revision Model Details} \label{app:reviser-detail}

The end-to-end \emph{reviser} training pipeline is visually shown in Figure \ref{fig:reviser-architecture}.

\section{Summary Training Details} \label{app:sum-train-detail}

For training, we use two abstractive models: BART \citep{lewis2019bart} and the encoder-decoder Longformer (LED) \citep{beltagy2020longformer} - a scaled up version of Bart to handle longer sequences via windowed local self-attention and a constant global attention. We fine-tune pre-trained checkpoints (\texttt{facebook/bart-base} and \texttt{allenai/led-base-16384} from the HuggingFace transformers library \citep{wolf-etal-2020-transformers}) for a maximum of 10,000 training steps, a maximum learning rate of $2e-5$ with a linear warmup of 200 steps, followed by linear decay to 0, and a batch size of 16. For Longformer, we use a maximum input length of 10,000. The maximum encoder length is 16,384 yet we could only fit 10,000 tokens onto a single 16GB V100 GPU. Training took approximately 1 day to complete 10 epochs on a single V100. For Bart, we use the maximum input length of 1,024. To handle longer input sequences, we rely on oracle filtering by taking the sentences with the largest average of ROUGE-1, ROUGE-2, and ROUGE-L F1 vis-a-vis the original reference. For models trained on revised data, oracle extraction is computed based on revised references during training yet the original during testing. This puts it a slight disadvantage to the other models which do not have this train-test mismatch. Yet, this mismatch only strengthens the empirical hypothesis considering the performance gains we see. During generation, we set beam search to $4$, use trigram blocking, and set the maximum output length to 1,024.

\begin{table*}[t]
\centering
\footnotesize 
\begin{tabular}{cc|ccccc}
\textbf{Reference} & \textbf{Quality} & \textbf{\# Training} & & & \multicolumn{2}{c}{\textbf{Entity-Level Source Overlap}} \\
\textbf{Version} & \textbf{Strategy} & \textbf{Examples} & \textbf{Avg. \# Tokens} & \textbf{Avg. \# Entities} & \textbf{Avg. Precision}& \textbf{Avg. Recall} \\ \hline
\textbf{Original} & Original & 45k & 370 & 50 & 60.4 & 26.8 \\
\hline
\multirow{2}{*}{\textbf{Filtered}} & No Admission & 5.7k & 420 & 64 & 91.5 & 32.7 \\
 & Unsupported & 6.0k & 381 & 55 & 95.7 & 28.9 \\  \hline
\multirow{3}{*}{\textbf{\makecell{Revised\\(Ours)}}} & Fully Extractive & 45k & 295 & 28 & 100 & 19.6 \\
 & Less Abstractive & 45k & 300 & 29 & 97.6 & 36.8 \\
 & More Abstractive & 45k & 272 & 26 & 96.8 & 35.0  \\ \hline
\end{tabular}
\caption{Training datasets obtained from the revision strategies according to size (number of references), average number of tokens and entities in the references, and entity-level overlap with sources (average precision and recall).} 
\label{tab:revision}
\end{table*}

\section{Statistics on Revised Training Datasets} \label{app:revisedtraining-data}

Table~\ref{tab:revision} describes statistics from original, filtered, and revised versions of the training data. As expected, the Filtered datasets (5.7k/6.0k) are smaller than the Original and Revised sets (45k), indicating the high number of noisy references. The Filtered references are more faithful than Original, as reflected by entity-level source precision (e.g., 91.5/95.7 vs 60.4). In comparison, the abstractively revised references are more concise than Original and Filtered at the token (e.g., 300/272 vs 370 and 420/381) and entity level (e.g., 29/26 vs 50 and 64/55), yet contain a larger fraction of the entities present in the source (36.8/35.0 versus 26.8 and 32.7/28.9). While the fully extractive references have perfect average precision (by construction), the abstractive revisions are close, 97.6/96.8, and contain almost twice as many relevant entities (36.8/35.0 vs 19.6).

\begin{table*}[t!]
\small
\centering
\begin{tabular}{c|cc|cccccc}
& \textbf{\makecell{Reference \\ Version}} & \textbf{\makecell{Quality \\ Strategy}} & \textbf{\makecell{Hallucination\\Rate (HR) $\downarrow$}} & \multicolumn{3}{c}{\textbf{\makecell{BERTScore\\P / R / F1 (BS) $\uparrow$}}} & \textbf{\makecell{Entail. \\$\uparrow$}} & \textbf{\makecell{Faithful-Adjusted \\ Recall (FaR) $\uparrow$}} \\ \hline 
\hline
\parbox[t]{1ex}{\multirow{9}{*}{\rotatebox[origin=c]{90}{\sc{\textbf{{Bart}}}}}} & \multirow{2}{*}{\textbf{Original}} & N/A & \textbf{38.9} & 81.3 & 69.2 & 74.7 & \textbf{43.6} & \textbf{47.7} \\ 
& & Control Halluc. & 40.3 & 81.7 & 69.2 & 74.8 & 43.2 & 46.6 \\ \cline{2-9}
& \multirow{4}{*}{\textbf{\makecell{Filtered \\ (Baselines)}}} & No Admission & 25.6 & 85.6 & 70.7 & 77.3 & 56.0 & 44.4 \\ 
& & Unsupported & \textbf{22.9} & 86.5 & 71.1 & 77.9 & \textbf{59.6} & \textbf{47.2} \\ 
& & Loss Truncation & 40.9 & 81.3 & 69.1 & 74.6 & \textbf{51.6} & 45.1 \\ 
& & Halluc. Ents & 37.6 & 82.3 & 69.1 & 75.0 & 48.6 & 46.4 \\ \cline{2-9} 
& \multirow{3}{*}{\textbf{\makecell{Revised\\(Ours)}}} & Fully Extractive & \textbf{9.1} & 92.3 & 72.8 & 81.2 & 72.1 & 52.1 \\ 
& & Less Abstractive & 7.3 & 91.8 & 72.9 & 81.1 & 72.5 & \textbf{56.3} \\ 
& & More Abstractive & \textbf{7.4} & 90.7 & 72.2 & 80.3 & 69.2 & 56.3  \\ \hline 
\end{tabular}
\caption{Summarization quality metrics across reference quality mitigation strategies (original, filtered, control, revised) used for training BART summarization models.}
\label{tab:results-bart}
\end{table*}

\section{Human Evaluation Setup} \label{app:human-eval}

As described in \citet{fabbri2021summeval}, we solicit summary feedback from an in-house clinical expert on 4 critical dimensions: Consistency, Relevance, Fluency, and Coherence. The annotator was provided the following guidance on each metric:

\begin{itemize}
    \item \textbf{Consistency}: The rating measures whether the facts in the summary are consistent with the facts in the original article. Consider whether the summary does reproduce all facts accurately and does not make up untrue information.
    \item \textbf{Relevance}: The rating measures how well the summary captures the key points of the article. Consider whether all and only the important aspects are contained in the summary.
    \item \textbf{Fluency}: The rating measures the quality of individual sentences: are they well-written and grammatically correct? Consider the quality of individual sentences.
    \item \textbf{Coherence}: The rating measures the quality of all sentences collectively: do they fit together and sound naturally?  Consider the quality of the summary as a whole.
\end{itemize}

The annotator was asked to assess summaries with independent rankings for each of the 4 metrics along a 1-5 Likert Scale. Given the complexity of the task and resource constraints, we sampled a set of 10 summaries from the test set. We first discard length outliers: the 10\% of examples with the fewest source tokens (insufficient input) as well as the 10\% of examples with most source tokens (too difficult to evaluate), before sampling at random. It took the expert 4 days to review 10 patient charts.

\section{BART Summarization Results} \label{app:bart-results}

For robustness, we evaluate our methods by fine-tuning from BART models, as well as Longformer Encoder-Decoder (LED) models. Table \ref{tab:results-bart} reveals similar findings for BART models as Table \ref{tab:results} revealed for Longformer. These findings demonstrate a degree of invariance to summary compression ratios, since BART only accepts 1,024 tokens, yet the maximum target output was 1,024 tokens for both.



\begin{figure*}[t]
\includegraphics[width=\linewidth]{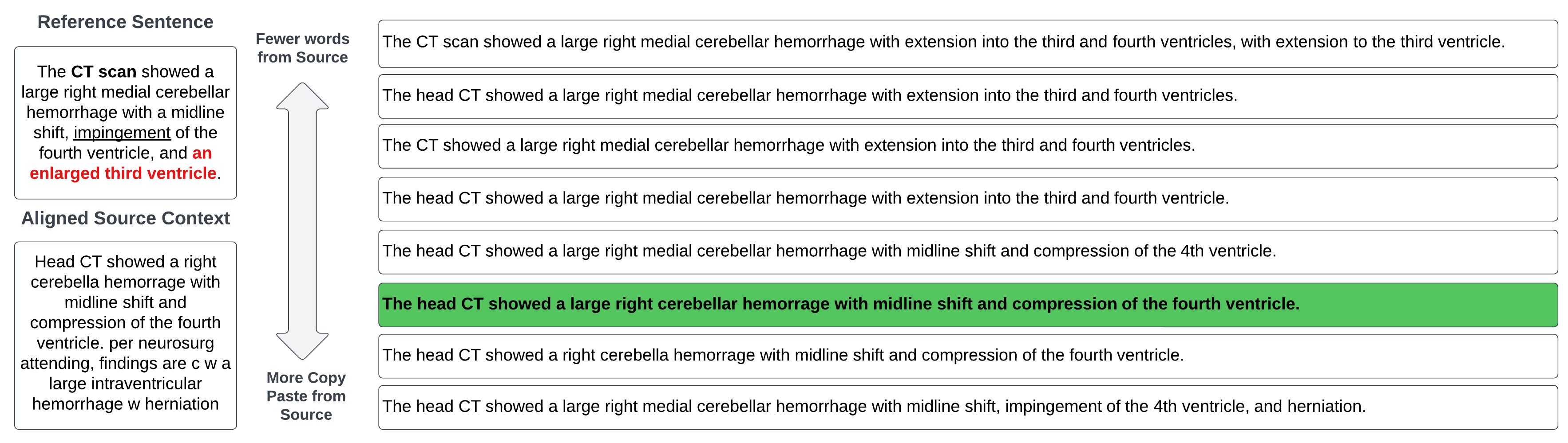}
\caption{Over-Generated Revisions from varying revision style codes. The \textcolor{Green}{highlighted} sentence is the one ultimately selected as the revision.}
\label{fig:over-generated}
\centering
\end{figure*}

\section{Qualitative Analysis} \label{app:qualitative}

We show an Abstractive example which received low assessed coherence (1) (other scores were all 3) to better understand its relative under-performance:

\begin{quote}
\small
A small right sub pectoral subcutaneous emphysema is mildly increased from the recent outside hospital exam. He was found to have a small right pneumothorax and multiple rib fractures on the right. A chest x-ray was obtained which showed a small anterior mediastinal hematoma likely tracking from the adjacent rib fractures. A CT of the chest was obtained to assess for progression of PTX contusions. There was no evidence of lung contusion or laceration on chest CT. He had no further episodes of bradycardia while in the Trauma ICU. A repeat chest CT was obtained on <date>.
\end{quote}


The Brief Hospital Course should roughly follow the same arc: presenting problem, investigation, diagnosis, treatment, (any complications and treatment), and, possibly, any follow-up plan. Deviation from this standard was penalized with lower coherence scores. The above summary does not follow this arc and the last sentence does not provide new information regarding the chest CT.

\paragraph{Reviser Outputs at Different Intensities.} In Figure \ref{fig:over-generated}, we show diverse outputs from the reviser model, with each output conditioned on a different extractiveness code ($source_{frac}$ from \S \ref{sec:reviser}).  We see a relatively smooth interpolation from abstractive to extractive.  At low $source_{frac}$ codes, revisions include hallucinations regarding the third ventricle, but ultimately, the reviser edits them out, and, its place, introduces something that is only mentioned in the aligned source context: herniation. The green highlight is the sentence ultimately chosen by the More Abstractive strategy for its blend of abstraction and faithfulness (groundedness).

\end{document}